\documentclass{midl} 

\usepackage{xcolor,colortbl}
\usepackage{mwe} 
\usepackage{multirow}

\usepackage{listings}
\usepackage{xcolor}

\definecolor{mygreen}{rgb}{0,0.6,0}
\definecolor{mygray}{rgb}{0.5,0.5,0.5}
\definecolor{mymauve}{rgb}{0.8,0,0}

\newcommand{\spheading}[2][4.5em]{
  \rotatebox{45}{\parbox{#1}{\raggedright #2}}}

\jmlrvolume{}
\jmlryear{}
\jmlrworkshop{}

\title[MultiMedEval: A Benchmark and a Toolkit for Evaluating Medical VLMs]{MultiMedEval: A Benchmark and a Toolkit for Evaluating Medical Vision-Language Models}

\midlauthor{\Name{Corentin Royer} \Email{corentin.royer@uzh.ch}\\
\Name{Bjoern Menze\midljointauthortext{Joint supervision
}} \Email{bjoern.menze@uzh.ch}\\
\Name{Anjany Sekuboyina\midlotherjointauthor} \Email{anjany.sekuboyina@uzh.ch}\\
\addr University of Zürich 
}

\begin{document}

\maketitle

\begin{abstract}
We introduce MultiMedEval, an open-source toolkit for fair and reproducible evaluation of large, medical vision-language models (VLM). MultiMedEval comprehensively assesses the models' performance on a broad array of six multi-modal tasks, conducted over 23 datasets, and spanning over 11 medical domains. The chosen tasks and performance metrics are based on their widespread adoption in the community and their diversity, ensuring a thorough evaluation of the model's overall generalizability. We open-source a Python toolkit (\small{\url{github.com/corentin-ryr/MultiMedEval}}) with a simple interface and setup process, enabling the evaluation of any VLM in just a few lines of code. Our goal is to simplify the intricate landscape of VLM evaluation, thus promoting fair and uniform benchmarking of future models.

\end{abstract}


\section{Introduction}

Large language models (LLM) and vision-language Models (VLM) are text generators capable of tackling a multitude of tasks based on textual or textual-and-visual prompts, \textit{e.g.}~question answering, machine translation, summarization, visual-question answering, image captioning, image classification, etc. Typically, assessing the performance of these models means evaluating them over a variety of tasks (mentioned above) on diverse datasets. This enables a reliable tracking of their progress and generalizability. General-purpose language models are therefore popularly benchmarked on toolkits such as OpenAI Evals, Huggingface LLM leaderboard \cite{open-llm-leaderboard}, and OpenVLM Leaderboard \cite{2023opencompass}. Such leaderboards offer a common platform for comparing open-access (Llama~2 \cite{touvron2023llama} and Flamingo \cite{alayrac2022flamingo}), and oftentimes, closed-source models (GPT-4V \cite{achiam2023gpt} and Gemini \cite{team2023gemini}) based on their performance.

Adapting VLMs to the medical domain proves challenging, primarily due to the domain-specific hurdles posed by proprietary datasets, fine-grained knowledge requirements, and the overall difficulty to generalize across medical domains and tasks. Despite these challenges, recent efforts culminated in truly capable \emph{medical} VLMs. For instance, LLaVA-Med \cite{li2023llavamed}, and PMC-VQA \cite{zhang2023pmcvqa} build VLM assistants for medical VQA, while MAIRA-1 \cite{hyland2023maira1} focuses on radiology report generation, specifically chest x-rays (CXR). RadFM \cite{wu2023generalist} proposes a versatile VLM, with a focus on radiology. Circumscribing the capabilities of the above-mentioned VLMs, MedPaLM~M \cite{tu2023towards} and BiomedGPT \cite{zhang2023biomedgpt} are pitched as \emph{generalist} models capable of performing a wider array of tasks such as image classification, text summarization, etc. 

Among these plethora of models, evaluation has been highly non-uniform. For instance, RadFM is evaluated on seven tasks while MedPaLM~M is evaluated on five. Even among the tasks that both of them have been evaluated on, discrepancies exist in terms of either the datasets (for report generation MedPALM is evaluated MIMI-CXR while RadFM is evaluated on three more datasets) or the metrics (BLEU and Recall in RadFM; BLEU and F1 in MedPALM). Similarly, LLAVA-Med reports six metrics, while RadFM reports 61 metrics; however, they share no common metrics, hindering a direct comparison of both approaches. Thus, there is a need for a unified benchmark, the lack of which has been consistently acknowledged \cite{wu2023generalist,li2023llavamed}. Our work aims to build such a unified benchmark. Closely related to our work are the works by \cite{wu2023generalist} and \cite{tu2023towards}. However, the benchmark of \cite{wu2023generalist} is specific to radiology-related tasks and doesn't include important domains such as general medicine. While the evaluation performed by \cite{tu2023towards} is holistic, it is closed-source, preventing a fair replication.

\begin{figure}[!t]
  \centering
  \includegraphics[width=\textwidth]{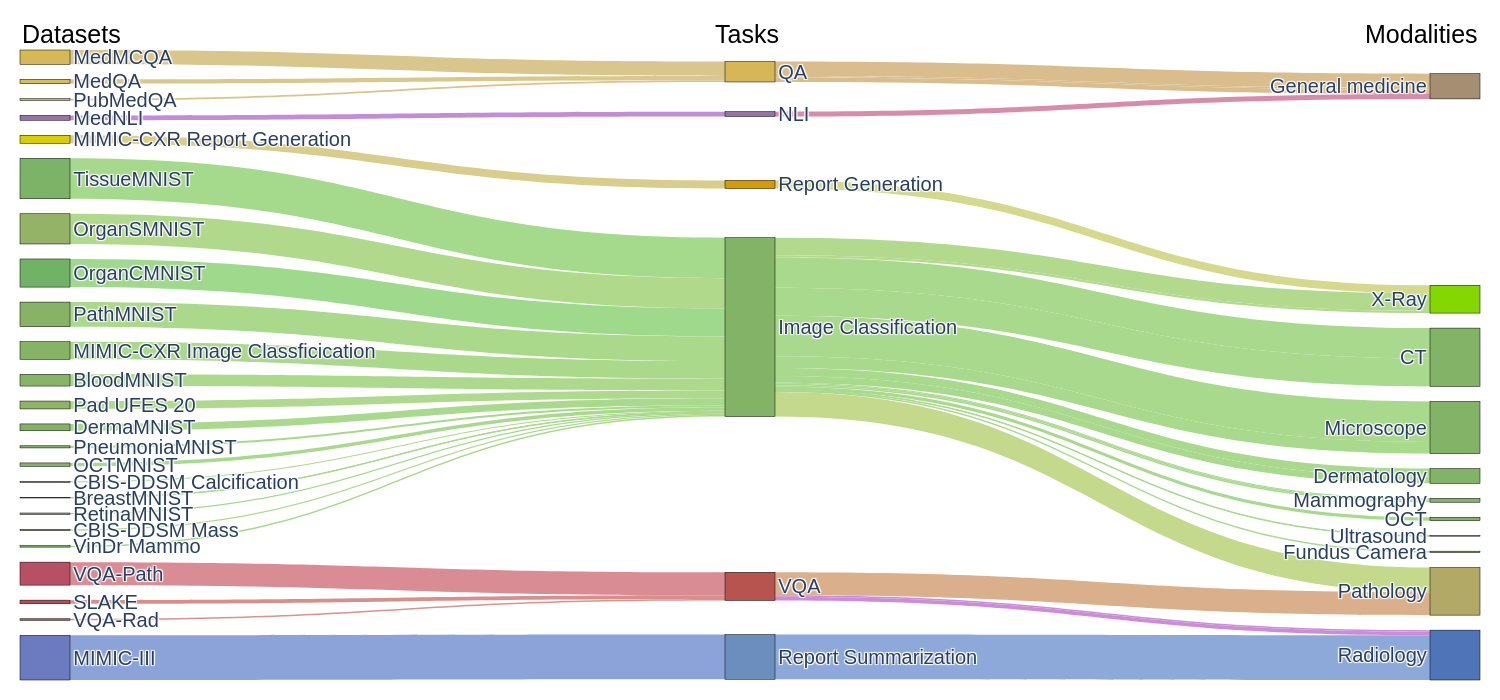}
  \caption{\small{The left column represents the datasets in MultiMedEval and their size. Each dataset is used in a task, represented in the middle. Finally, we represent the share of each modality in the tool and the modality composition of each task.}}
  \label{fig:sankey}
\end{figure}

Summarizing the topic of the evaluation of medical VLMs, we identify three key issues that make evaluation difficult and eventually slow down medical VLM research. First, as stated above, models are benchmarked using different metrics, methodologies, and tasks preventing fair comparison to other models. Second, the scope of generalist models is so wide that every new model is required to benchmark against all the prior work. However, re-implementing the metrics and recomputing benchmarks is time-consuming when models are open-access and impossible for closed-source models. Third, even if one plans to re-implement the benchmark, the evaluation pipelines (from data to metrics) are complex and cumbersome. Addressing these issues, \textbf{we introduce MultiMedEval, an open-source, Python-based evaluation toolkit for medical VLMs}. Our toolkit is designed to be user-friendly, offering reproducible evaluation capabilities across six distinct medical tasks implemented on 23 datasets spanning over 11 medical image and text modalities. Our benchmark encompasses a broader spectrum of medical domains than any model mentioned above. Additionally, we reimplemented two open-source models (RadFM and LLAVA-Med) and compared them with closed models according to the metrics reported therein. Our comprehensive comparison serves as a foundational baseline for future medical VLM research, with future evaluations seamlessly unified into this benchmark through MultiMedEval, resulting in its continued growth.

\section{Tasks and Evaluations}
\label{sec:tasks_and_evals}

The typical VLM is designed to take an interleaved image-text prompt as input and generate a textual response as output. For example, the prompt \texttt{\small{<image> What is the modality of this image?}} might generate a response \texttt{\small{This image shows an MRI of the brain}}. Any evaluation primarily focuses on the model's ability to respond to a variety of question prompts, which in turn defines the variety of \emph{tasks} the model can perform. For a comprehensive evaluation of the VLM's capabilities, we evaluate it on six tasks: image classification, question answering (QA), visual QA, report summarization, report generation, and natural language inference (NLI). In every task, we detail the prompt design, datasets, and performance metrics that MultiMedEval employs during evaluation. We let the user decide if they want to do few-shot inference (\textit{i.e.}~prepend the prompt with examples of prompts and responses) or zero-shot inference. We use the official split for every dataset except MIMIC-III and Pad-UFES-20 which do not have an official split. We propose such a split for Pad-UFES 20 where we use 20\% of the dataset for testing and we use the split proposed by \cite{delbrouck-etal-2022-vilmedic} for MIMIC-III. Fig.~\ref{fig:sankey} gives an overview of the dataset, task, and modality distribution in MultiMedEval. For every task, example prompts for all the datasets are listed in the appendix.

\noindent
\paragraph{Multi-class and multi-label image classification.}

We use a total of 15 datasets spanning nine modalities (\textit{c.f.}~\tableref{tab:imageClassificationModalities}). MIMIC-CXR is a multi-label classification task while the others are multi-class classification. The input prompt for image classification is constructed by the image followed by the classes and then a question, \textit{e.g.} for OrganMNIST the full prompt will be: \texttt{\small{<img> Options:1:bladder  2:femur-left ... 11: spleen Which options correspond to the image?}}. 

For all datasets, except MIMIC-CXR, the model's predicted answer is determined by calculating the BLEU score between the model's response and each class, selecting the class with the highest score. For the MIMIC-CXR dataset, we use CheXBert \cite{smit2020chexbert} labeler on the response. CheXBert is a report-labeling tool that extracts 14 conditions of which we keep five conditions (Atelectasis, Cardiomegaly, Consolidation, Edema, and Pleural Effusion) for computing the metrics. Once the classes are extracted, we report the classification performance using macro F1, macro AUROC, and macro accuracy. 

\begin{table}[t]
\tiny
\centering
\caption{\small List of the image classification datasets and the different modalities they cover.}
\label{tab:imageClassificationModalities}
    \begin{tabular}{c c c c}
    \textbf{Modality}            & \textbf{Dataset name}                                    & \textbf{Classes} & \textbf{Size}   \\ \hline
    \multirow{2}{*}{CT (Radiology)}          & OrganSMNIST \cite{xu2019efficient,bilic2023liver} & 11               &  8827          \\  
                                 & OrganCMNIST \cite{xu2019efficient,bilic2023liver} & 11               &  8216          \\ \hline
    \multirow{2}{*}{Dermatology} & Pad-UFES 20 \cite{pacheco2020pad}                 & 7                &  2298         \\  
                                 & DermaMNIST \cite{ham10000,codella2019skin}       & 7                &  2005     
                                 \\ \hline
    Fundus Camera                & RetinaMNIST \cite{liu2022deepdrid}                       & 5                &  400          \\ \hline
    \multirow{3}{*}{Mammography} & VinDr Mammo \cite{nguyen2023vindr}                       & 5                &  429          \\  
                                 & CBIS-DDSM Mass \cite{lee2017curated}                     & 3                &  378          \\  
                                 & CBIS-DDSM Calcification \cite{lee2017curated}            & 3                &  326          \\ \hline
    \multirow{2}{*}{Microscope}  & TissueMNIST \cite{ljosa2012annotated}                    & 8                &  11820        \\ 
                                 & BloodMNIST \cite{acevedo2020dataset}                     & 8                &  3421         \\ \hline
    OCT                          & OCTMNIST \cite{kermany2018identifying}                   & 4                &  1000         \\ \hline
    Pathology                    & PathMNIST \cite{kather2019predicting}                    & 9                &  7180         \\ \hline
    Ultrasound                   & BreastMNIST \cite{al2020dataset}                         & 2                &  7180          \\ \hline
    \multirow{2}{*}{X-Ray (Radiology)}       & PneumoniaMNIST \cite{kermany2018identifying}             & 2                &  7180          \\  
                                 & MIMIC Image Classification \cite{johnson2019mimic}       & 5                &  5159 \\
    \end{tabular}
\end{table}

\noindent

\begin{table}[t]
  \tiny
  \begin{minipage}{0.45\textwidth}
    \centering
    \caption{\small{List of the QA datasets and the different modalities they cover.}}
    \label{tab:QAModalities}
    \begin{tabular}{c c c}
       \textbf{Modality}        & \textbf{Dataset name}              & \textbf{Size}  \\ \hline
    \multirow{3}{*}{General Medicine}         & MedQA \cite{lau2018dataset}        &  1273          \\ 
             & MedMCQA \cite{he2020pathological}  &  4183          \\ 
             & PubMedQA \cite{liu2021slake}       &  500           \\ 
    \end{tabular}
    
  \end{minipage}
  \hfill 
  \begin{minipage}{0.45\textwidth}
    \centering
    \caption{\small{List of the VQA datasets and the different modalities they cover.}}
    \label{tab:VQAModalities}
    \begin{tabular}{c c c}
       \textbf{Modality}        & \textbf{Dataset name}              & \textbf{Size}   \\ \hline
    \multirow{2}{*}{Radiology}                & VQA-Rad \cite{lau2018dataset}      &  451            \\ 
                    & SLAKE \cite{liu2021slake}          &  1061           \\ \hline
            Pathology        & Path-VQA \cite{he2020pathological} &  6719           \\ 
    \end{tabular}
   
  \end{minipage}
  
\end{table}

\paragraph{Question answering.}

For QA, we evaluate on three datasets (\textit{c.f.}~Table~\ref{tab:QAModalities}): MedQA and MedMCQA consist of multi-choice questions (MCQ) and PubMedQA consists of close-ended questions (yes-no questions). For the MCQs, the model is prompted with the question followed by all the options and finally the phrase \texttt{\small{What is the correct answer?}}. For PubMedQA, the question is prepended with \texttt{\small{Answer the question with yes, no, or maybe}}. 

 To determine the model's predicted answer from its response, we utilize the BLEU metric \cite{papineni2002bleu} to compare the predicted answer to each option, selecting the one with the highest BLEU score. For PubMedQA, we check if the answer contains the words \texttt{\small{yes}}, \texttt{\small{no}}, or \texttt{\small{maybe}}. Once the model's answers have been generated, we report answering performance using accuracy.

\paragraph{Visual question answering.}
For evaluating the performance of a VLM on VQA, we use three datasets (\textit{c.f.}~Table~\ref{tab:VQAModalities}) containing a mix of open-ended and close-ended questions: Path-VQA, SLAKE, and VQA-Rad. The prompt for VQA is constructed by concatenating the image and the question, \textit{e.g.} \texttt{\small{<img> What is the main organ in the image?}}. Similar to QA, the close-ended questions are prepended with \texttt{\small{Answer the question with yes or no}}.

Since there are no MCQs in VQA, the evaluation differs from QA. Specifically, the correct answer and the predicted one are tokenized and the resulting sets are used to compute precision and recall. We also differentiate between close-ended and open-ended questions to report close-ended accuracy, open-ended accuracy, and open-ended recall \cite{nguyen2019overcoming}. Additionally, we also report overall recall and F1 score. Finally, we compute the BLEU score from the non-tokenized texts \cite{wu2023generalist,li2023llavamed}. To calculate the accuracy, close-ended questions are considered correct if their recall is at least 0.5, while open-ended questions require a recall of at least 0.75.

\paragraph{Report generation.}
We include the MIMIC-CXR dataset \cite{johnson2019mimic} containing de-identified radiology reports with the associated CXRs to evaluate report-generation capabilities. The task is to generate the \emph{findings} section of the report based on the radiology images. The input prompt for the model is constructed with the image (or multiple images pertaining to one case) followed by a sentence asking for the report, as in, \texttt{\small{<img> <img> Please caption this scan with findings and impressions}}. 

Following common practices, the generated reports are evaluated using n-gram-based methods: ROUGE-L, BLEU-1, BLEU-4, and METEOR \cite{banerjee2005meteor}. Additionally, we compute F1-RadGraph, CheXBert vector similarity, F1-bertscore, and RadCliQ to capture the subtleties of radiological language. F1-RadGraph is the F1 score between the entities extracted from the reference report and generated one using RadGraph \cite{jain2021radgraph}. F1-BertScore employs CheXBert to label the reference and generated reports, as introduced in the image classification task above. CheXBert vector similarity \cite{yu2023evaluating} computes the cosine similarity between the embedded reference and generated reports. Lastly, RadCliQ \cite{yu2023evaluating} is a composite metric composed of the previous four metrics, said to closely match the practitioners' feedback on report quality.

\paragraph{Report summarization.}
\label{reportSummarization}

We evaluate report summarization on MIMIC-III \cite{johnson2016mimic}. Following \cite{van2023radadapt}, the VLM has to generate the \emph{impressions} section of a radiology report based on the \emph{findings} section. So, the model is prompted with the free-text findings (\textit{e.g.} \texttt{\small{\hyphenchar\font=\string"7F Intracranial vessels are normal... there is mild ventriculomegal... the subarachnoid hemorrhage noted in the right sylvian fissure}}) followed by the task prompt, \texttt{\small{Summarize the findings}}. We use the same metric as for the report generation task to compare the ground truth impressions to the generated summary.

\paragraph{Natural language inference.}

Gauging the logical reasoning capabilities of the VLM, Natural language inference (NLI) involves the categorization of pairs of sentences into three classes: contradiction, entailment, or neutral, effectively resembling a 3-class classification task. We evaluate NLI using MedNLI \cite{romanov2018lessons}, a dataset consisting of pairs of medical statements. We prompt the model with the two sentences from a pair followed by a question asking to classify the logical relationship between them, as in \texttt{\small{\hyphenchar\font=\string"7F <sentence-1> <sentence-2> Determine the logical relationship between these two sentences.}}

To extract the VLM's predicted answer from its response, we check the presence of either of the three terms (contradiction, entailment, or neutral). For the answer to be valid, only one of the three classes must be present; if none of the three or several of them are present, the answer is deemed invalid. The performance of the model is then reported using accuracy.

\section{MultiMedEval Setup and Utilization}

Ideally, once a VLM is developed, the entire suite of evaluations mentioned above needs to be conducted. Typically, this involves downloading the datasets, implementing the data pre-processing, implementing the computation of the metrics, running inference through the VLM, and finally recording the performance. We design MultiMedEval to abstract this entire pipeline, only exposing APIs for setting up the data and for evaluating on them. In this section, we briefly describe the usage of MultiMedEval and strongly encourage the reader to peruse the official documentation (currently hosted at \url{github.com/corentin-ryr/MultiMedEval}).

\noindent
\paragraph{Parameters.} MultiMedEval exposes two parameter classes, \lstinline{SetupParams} and \lstinline{EvalParams}. The setup parameters control the data download. For the datasets hosted on PhysioNet, the setup process also requires appropriate credentials. The evaluation parameters control the evaluation configuration such as \lstinline{batch_size}, \lstinline{device} (GPU-id), etc.

\noindent
\paragraph{Batcher.} The only code that the user needs to implement is a Callable, \lstinline{batcher}, which wraps around the user's VLM inference module. Every call to \lstinline{batcher} takes, as input, a batch of conversation prompts and returns, as output, the decoded model responses. The input prompts are constructed according to HuggingFace's conversation style.

\noindent
\paragraph{Evaluation.}
Once the parameters and the \lstinline{batcher} are ready, the evaluation can be run using the \lstinline{eval} API, which takes as input the list of datasets that the user wants to benchmark on, along with the \lstinline{batcher} and the evaluation parameters. The results of the evaluation are saved as a JSON file. Below, we provide the pseudocode for one such evaluation:

\begin{lstlisting}[language=Python]
from multimedeval import MultiMedEval, SetupParams, EvalParams

# Implementing the batcher for the user's specific model and returning text answers
def exampleBatcher(prompts:list[tuple]) -> list[str]:
    return [model.generate(prompt) for prompt in prompts]

engine = MultiMedEval()
# Running the setup only for MedQA
setupParams = SetupParams(MedQA_dir="data/")
engine.setup(setupParams)

# Running the evaluation on the exampleBatcher
evalParams = EvalParams(batch_size=32)
engine.eval(["MedQA"], exampleBatcher, evalParams)
\end{lstlisting}

\begin{table}[t]
    \tiny
    \centering
    \caption{Performance of baseline VLMs on MultiMedEval's tasks. Brighter the cell, better the performance. Grey values indicate metrics that the model was not evaluated with. Table continued in Table~\ref{tab:allResults2}.\label{tab:allResults1}}%
\begin{tabular}{p{0.1\linewidth}p{0.15\linewidth}p{0.05\linewidth}p{0.05\linewidth}p{0.05\linewidth}p{0.05\linewidth}p{0.05\linewidth}p{0.05\linewidth}p{0.05\linewidth}}
 Dataset                                                      & Metric              & \spheading{RadFM}                                                                 & \spheading{RaFM (reported)}   & \spheading{LLaVA-Med}                                                             & \spheading{LLaVA-Med (reported)}   & \spheading{Med-PaLM M (reported)}                                                   & \spheading{Maira-1 (reported)}   & \spheading{Biomedgpt (reported)}                                                    \\
\hline
 \multicolumn{9}{c}{\cellcolor{black!35}Image classification}        \\
 \multirow{3}{\linewidth}{MIMIC-CXR Image classification}     & Macro-AUC$\uparrow$ & \cellcolor[rgb]{0.3607843137254902,0.615686274509804,0.3607843137254902} 0.536    & \cellcolor{black!25} -        & \cellcolor[rgb]{0.39215686274509803,0.5882352941176471,0.39215686274509803} 0.470 & \cellcolor{black!25} -             & \cellcolor[rgb]{0.1607843137254902,0.8274509803921568,0.1607843137254902} 0.791     & \cellcolor{black!25} -           & \cellcolor{black!25} -                                                              \\
                                                              & Accuracy$\uparrow$  & \cellcolor[rgb]{0.10588235294117647,0.8862745098039215,0.10588235294117647} 0.728 & \cellcolor{black!25} -        & \cellcolor[rgb]{0.12941176470588237,0.8627450980392157,0.12941176470588237} 0.668 & \cellcolor{black!25} -             & \cellcolor{black!25} -                                                              & \cellcolor{black!25} -           & \cellcolor[rgb]{0.0392156862745098,0.9568627450980393,0.0392156862745098} 0.897     \\
\end{tabular}
\end{table}

\begin{table}[t]
    \tiny
    \centering
    \caption{Continuation of the VLM benchmark from Table~\ref{tab:allResults1}.\label{tab:allResults2}}%
    \begin{tabular}{p{0.1\linewidth}p{0.15\linewidth}p{0.05\linewidth}p{0.05\linewidth}p{0.05\linewidth}p{0.05\linewidth}p{0.05\linewidth}p{0.05\linewidth}p{0.05\linewidth}}
 Dataset                                                      & Metric                      & \spheading{RadFM}                                                                 & \spheading{RaFM (reported)}                                                       & \spheading{LLaVA-Med}                                                             & \spheading{LLaVA-Med (reported)}                                                  & \spheading{Med-PaLM M (reported)}                                                   & \spheading{Maira-1 (reported)}                                                    & \spheading{Biomedgpt (reported)}                                                    \\
\hline
 \multicolumn{9}{c}{\cellcolor{black!35}QA}                        \\
 \multirow{1}{\linewidth}{MedQA}                              & Accuracy$\uparrow$          & \cellcolor[rgb]{0.30196078431372547,0.6823529411764706,0.30196078431372547} 0.230 & \cellcolor{black!25} -                                                            & \cellcolor[rgb]{0.29411764705882354,0.6862745098039216,0.29411764705882354} 0.241 & \cellcolor{black!25} -                                                            & \cellcolor[rgb]{0.11764705882352941,0.8745098039215686,0.11764705882352941} 0.697   & \cellcolor{black!25} -                                                            & \cellcolor{black!25} -                                                              \\
\end{tabular}
\end{table}

\section{Baselines}
As previously indicated, MultiMedEval's purpose is to enable a comprehensive assessment of any VLM. To demonstrate this, we benchmark two recent, publicly-available models, RadFM \cite{wu2023generalist} and LLaVA-Med \cite{li2023llavamed}. In this benchmark, we also include the performance reported by two closed models, MedPALM~M \cite{tu2023towards} and Maira-1 \cite{hyland2023maira1}, as well as one very recent public model, BiomedGPT \cite{zhang2023biomedgpt}. In Tables~\ref{tab:allResults1}~and~\ref{tab:allResults2}, we report the complete picture of the model performances across six tasks, 23 datasets, and 81 metrics. The performance is grouped by tasks and color-coded in green. The brighter the color, the superior the performance (\textit{i.e.}~highest accuracy or lowest RadCliQ score is the brightest). A gray cell indicates that the performance was never reported.

Owing to a holistic picture provided by MultiMedEval, we make five crucial observations: First, there is not a single task or metric that every medical VLM, to date, has been evaluated on. This speaks to the non-uniformity in the existing evaluation regimes. Second, closed models such as MAIRA-1 (on report generation) and MedPaLM~M (on QA, image classification, etc.) show superior performance compared to open-source models. Third, we can see an improvement in VLMs' performance on QA and image classification. In both cases, MedPaLM~M outperforms the others by a significant margin. Fourth, the most recent open-source model (BiomedGPT) seems to be very promising. On metrics reported by both MedPaLM~M and BiomedGPT, the latter shows a competitive performance. At image classification (Macro-F1 on CBIS-DDSM), BiomedGPT even outperforms MedPaLM~M, showing an encouraging prospect for open-source models. Fifth and finally, the number of empty cells clearly showcases the need for a standardized evaluation protocol for medical VLMs as well as an easy-to-use toolkit that performs the evaluation so that researchers can cater more focus on the development of the models.

\section{Conclusion}

Medical vision-language models are just gathering momentum, they already show interesting generalizable capabilities, and their capabilities are bound to expand. Therefore, this is an opportune moment to establish a standard evaluation protocol based on community consensus. Addressing this, we presented MultiMedEval, a Python toolkit to comprehensively assess the performance of any VLM model on multiple medical tasks. Using this, we benchmarked RadFM and LLaVVa-Med and compared their results to the reported performances of state-of-the-art medical VLMs.  
\noindent
\paragraph{Future work.} MultiMedEval will be released to the community and will be actively maintained by adding new tasks, metrics, and datasets. To this end, we will work with open-source medical imaging libraries such as MONAI \cite{monai} and MLCommons \cite{mlcommons} to increase community adoption.

\bibliography{bibliography}

@misc{wu2023generalist,
      title={Towards Generalist Foundation Model for Radiology by Leveraging Web-scale 2D\&3D Medical Data}, 
      author={Chaoyi Wu and Xiaoman Zhang and Ya Zhang and Yanfeng Wang and Weidi Xie},
      year={2023},
      eprint={2308.02463},
      archivePrefix={arXiv},
      primaryClass={cs.CV}
}

@misc{li2023llavamed,
      title={LLaVA-Med: Training a Large Language-and-Vision Assistant for Biomedicine in One Day}, 
      author={Chunyuan Li and Cliff Wong and Sheng Zhang and Naoto Usuyama and Haotian Liu and Jianwei Yang and Tristan Naumann and Hoifung Poon and Jianfeng Gao},
      year={2023},
      eprint={2306.00890},
      archivePrefix={arXiv},
      primaryClass={cs.CV}
}

@misc{zhang2023pmcvqa,
      title={PMC-VQA: Visual Instruction Tuning for Medical Visual Question Answering}, 
      author={Xiaoman Zhang and Chaoyi Wu and Ziheng Zhao and Weixiong Lin and Ya Zhang and Yanfeng Wang and Weidi Xie},
      year={2023},
      eprint={2305.10415},
      archivePrefix={arXiv},
      primaryClass={cs.CV}
}

@misc{hyland2023maira1,
      title={MAIRA-1: A specialised large multimodal model for radiology report generation}, 
      author={Stephanie L. Hyland and Shruthi Bannur and Kenza Bouzid and Daniel C. Castro and Mercy Ranjit and Anton Schwaighofer and Fernando Pérez-García and Valentina Salvatelli and Shaury Srivastav and Anja Thieme and Noel Codella and Matthew P. Lungren and Maria Teodora Wetscherek and Ozan Oktay and Javier Alvarez-Valle},
      year={2023},
      eprint={2311.13668},
      archivePrefix={arXiv},
      primaryClass={cs.CL}
}

@article{tu2023towards,
  title={Towards generalist biomedical ai},
  author={Tu, Tao and Azizi, Shekoofeh and Driess, Danny and Schaekermann, Mike and Amin, Mohamed and Chang, Pi-Chuan and Carroll, Andrew and Lau, Chuck and Tanno, Ryutaro and Ktena, Ira and others},
  journal={arXiv preprint arXiv:2307.14334},
  year={2023}
}

@article{lau2018dataset,
  title={A dataset of clinically generated visual questions and answers about radiology images},
  author={Lau, Jason J and Gayen, Soumya and Ben Abacha, Asma and Demner-Fushman, Dina},
  journal={Scientific data},
  volume={5},
  number={1},
  pages={1--10},
  year={2018},
  publisher={Nature Publishing Group}
}

@article{he2020pathological,
  title={Pathological visual question answering},
  author={He, Xuehai and Cai, Zhuo and Wei, Wenlan and Zhang, Yichen and Mou, Luntian and Xing, Eric and Xie, Pengtao},
  journal={arXiv preprint arXiv:2010.12435},
  year={2020}
}

@inproceedings{liu2021slake,
  title={Slake: A semantically-labeled knowledge-enhanced dataset for medical visual question answering},
  author={Liu, Bo and Zhan, Li-Ming and Xu, Li and Ma, Lin and Yang, Yan and Wu, Xiao-Ming},
  booktitle={2021 IEEE 18th International Symposium on Biomedical Imaging (ISBI)},
  pages={1650--1654},
  year={2021},
  organization={IEEE}
}

@inproceedings{nguyen2019overcoming,
  title={Overcoming data limitation in medical visual question answering},
  author={Nguyen, Binh D and Do, Thanh-Toan and Nguyen, Binh X and Do, Tuong and Tjiputra, Erman and Tran, Quang D},
  booktitle={Medical Image Computing and Computer Assisted Intervention--MICCAI 2019: 22nd International Conference, Shenzhen, China, October 13--17, 2019, Proceedings, Part IV 22},
  pages={522--530},
  year={2019},
  organization={Springer}
}

@misc{smit2020chexbert,
      title={CheXbert: Combining Automatic Labelers and Expert Annotations for Accurate Radiology Report Labeling Using BERT}, 
      author={Akshay Smit and Saahil Jain and Pranav Rajpurkar and Anuj Pareek and Andrew Y. Ng and Matthew P. Lungren},
      year={2020},
      eprint={2004.09167},
      archivePrefix={arXiv},
      primaryClass={cs.CL}
}

@article{johnson2016mimic,
  title={MIMIC-III, a freely accessible critical care database},
  author={Johnson, Alistair EW and Pollard, Tom J and Shen, Lu and Lehman, Li-wei H and Feng, Mengling and Ghassemi, Mohammad and Moody, Benjamin and Szolovits, Peter and Anthony Celi, Leo and Mark, Roger G},
  journal={Scientific data},
  volume={3},
  number={1},
  pages={1--9},
  year={2016},
  publisher={Nature Publishing Group}
}

@article{jain2021radgraph,
  title={Radgraph: Extracting clinical entities and relations from radiology reports},
  author={Jain, Saahil and Agrawal, Ashwin and Saporta, Adriel and Truong, Steven QH and Duong, Du Nguyen and Bui, Tan and Chambon, Pierre and Zhang, Yuhao and Lungren, Matthew P and Ng, Andrew Y and others},
  journal={arXiv preprint arXiv:2106.14463},
  year={2021}
}

@article{van2023radadapt,
  title={RadAdapt: Radiology Report Summarization via Lightweight Domain Adaptation of Large Language Models},
  author={Van Veen, Dave and Van Uden, Cara and Attias, Maayane and Pareek, Anuj and Bluethgen, Christian and Polacin, Malgorzata and Chiu, Wah and Delbrouck, Jean-Benoit and Chaves, Juan Manuel Zambrano and Langlotz, Curtis P and others},
  journal={arXiv preprint arXiv:2305.01146},
  year={2023}
}

@article{yu2023evaluating,
  title={Evaluating progress in automatic chest x-ray radiology report generation},
  author={Yu, Feiyang and Endo, Mark and Krishnan, Rayan and Pan, Ian and Tsai, Andy and Reis, Eduardo Pontes and Fonseca, Eduardo Kaiser Ururahy Nunes and Lee, Henrique Min Ho and Abad, Zahra Shakeri Hossein and Ng, Andrew Y and others},
  journal={Patterns},
  volume={4},
  number={9},
  year={2023},
  publisher={Elsevier}
}

@article{johnson2019mimic,
  title={MIMIC-CXR, a de-identified publicly available database of chest radiographs with free-text reports},
  author={Johnson, Alistair EW and Pollard, Tom J and Berkowitz, Seth J and Greenbaum, Nathaniel R and Lungren, Matthew P and Deng, Chih-ying and Mark, Roger G and Horng, Steven},
  journal={Scientific data},
  volume={6},
  number={1},
  pages={317},
  year={2019},
  publisher={Nature Publishing Group UK London}
}

@article{xu2019efficient,
  title={Efficient multiple organ localization in CT image using 3D region proposal network},
  author={Xu, Xuanang and Zhou, Fugen and Liu, Bo and Fu, Dongshan and Bai, Xiangzhi},
  journal={IEEE transactions on medical imaging},
  volume={38},
  number={8},
  pages={1885--1898},
  year={2019},
  publisher={IEEE}
}

@article{bilic2023liver,
  title={The liver tumor segmentation benchmark (lits)},
  author={Bilic, Patrick and Christ, Patrick and Li, Hongwei Bran and Vorontsov, Eugene and Ben-Cohen, Avi and Kaissis, Georgios and Szeskin, Adi and Jacobs, Colin and Mamani, Gabriel Efrain Humpire and Chartrand, Gabriel and others},
  journal={Medical Image Analysis},
  volume={84},
  pages={102680},
  year={2023},
  publisher={Elsevier}
}

@article{pacheco2020pad,
  title={PAD-UFES-20: A skin lesion dataset composed of patient data and clinical images collected from smartphones},
  author={Pacheco, Andre GC and Lima, Gustavo R and Salomao, Amanda S and Krohling, Breno and Biral, Igor P and de Angelo, Gabriel G and Alves Jr, F{\'a}bio CR and Esgario, Jos{\'e} GM and Simora, Alana C and Castro, Pedro BC and others},
  journal={Data in brief},
  volume={32},
  pages={106221},
  year={2020},
  publisher={Elsevier}
}

@article{ham10000,
  title={The HAM10000 dataset, a large collection of multi-source dermatoscopic images of common pigmented skin lesions},
  author={Tschandl, Philipp and Rosendahl, Cliff and Kittler, Harald},
  journal={Scientific data},
  volume={5},
  number={1},
  pages={1--9},
  year={2018},
  publisher={Nature Publishing Group}
}

@article{codella2019skin,
  title={Skin lesion analysis toward melanoma detection 2018: A challenge hosted by the international skin imaging collaboration (isic)},
  author={Codella, Noel and Rotemberg, Veronica and Tschandl, Philipp and Celebi, M Emre and Dusza, Stephen and Gutman, David and Helba, Brian and Kalloo, Aadi and Liopyris, Konstantinos and Marchetti, Michael and others},
  journal={arXiv preprint arXiv:1902.03368},
  year={2019}
}

@article{liu2022deepdrid,
  title={Deepdrid: Diabetic retinopathy—grading and image quality estimation challenge},
  author={Liu, Ruhan and Wang, Xiangning and Wu, Qiang and Dai, Ling and Fang, Xi and Yan, Tao and Son, Jaemin and Tang, Shiqi and Li, Jiang and Gao, Zijian and others},
  journal={Patterns},
  volume={3},
  number={6},
  year={2022},
  publisher={Elsevier}
}

@article{nguyen2023vindr,
  title={VinDr-Mammo: A large-scale benchmark dataset for computer-aided diagnosis in full-field digital mammography},
  author={Nguyen, Hieu T and Nguyen, Ha Q and Pham, Hieu H and Lam, Khanh and Le, Linh T and Dao, Minh and Vu, Van},
  journal={Scientific Data},
  volume={10},
  number={1},
  pages={277},
  year={2023},
  publisher={Nature Publishing Group UK London}
}

@article{lee2017curated,
  title={A curated mammography data set for use in computer-aided detection and diagnosis research},
  author={Lee, Rebecca Sawyer and Gimenez, Francisco and Hoogi, Assaf and Miyake, Kanae Kawai and Gorovoy, Mia and Rubin, Daniel L},
  journal={Scientific data},
  volume={4},
  number={1},
  pages={1--9},
  year={2017},
  publisher={Nature Publishing Group}
}

@article{ljosa2012annotated,
  title={Annotated high-throughput microscopy image sets for validation.},
  author={Ljosa, Vebjorn and Sokolnicki, Katherine L and Carpenter, Anne E},
  journal={Nature methods},
  volume={9},
  number={7},
  pages={637--637},
  year={2012}
}

@article{acevedo2020dataset,
  title={A dataset of microscopic peripheral blood cell images for development of automatic recognition systems},
  author={Acevedo, Andrea and Merino, Anna and Alf{\'e}rez, Santiago and Molina, {\'A}ngel and Bold{\'u}, Laura and Rodellar, Jos{\'e}},
  journal={Data in brief},
  volume={30},
  year={2020},
  publisher={Elsevier}
}

@article{kermany2018identifying,
  title={Identifying medical diagnoses and treatable diseases by image-based deep learning},
  author={Kermany, Daniel S and Goldbaum, Michael and Cai, Wenjia and Valentim, Carolina CS and Liang, Huiying and Baxter, Sally L and McKeown, Alex and Yang, Ge and Wu, Xiaokang and Yan, Fangbing and others},
  journal={cell},
  volume={172},
  number={5},
  pages={1122--1131},
  year={2018},
  publisher={Elsevier}
}

@article{kather2019predicting,
  title={Predicting survival from colorectal cancer histology slides using deep learning: A retrospective multicenter study},
  author={Kather, Jakob Nikolas and Krisam, Johannes and Charoentong, Pornpimol and Luedde, Tom and Herpel, Esther and Weis, Cleo-Aron and Gaiser, Timo and Marx, Alexander and Valous, Nektarios A and Ferber, Dyke and others},
  journal={PLoS medicine},
  volume={16},
  number={1},
  pages={e1002730},
  year={2019},
  publisher={Public Library of Science}
}

@article{al2020dataset,
  title={Dataset of breast ultrasound images},
  author={Al-Dhabyani, Walid and Gomaa, Mohammed and Khaled, Hussien and Fahmy, Aly},
  journal={Data in brief},
  volume={28},
  pages={104863},
  year={2020},
  publisher={Elsevier}
}

@misc{open-llm-leaderboard,
  author = {Edward Beeching and Clémentine Fourrier and Nathan Habib and Sheon Han and Nathan Lambert and Nazneen Rajani and Omar Sanseviero and Lewis Tunstall and Thomas Wolf},
  title = {Open LLM Leaderboard},
  year = {2023},
  publisher = {Hugging Face},
  howpublished = "\url{https://huggingface.co/spaces/HuggingFaceH4/open_llm_leaderboard}"
}

@misc{2023opencompass,
    title={OpenCompass: A Universal Evaluation Platform for Foundation Models},
    author={OpenCompass Contributors},
    howpublished = {\url{https://github.com/open-compass/opencompass}},
    year={2023}
}

@article{zhang2023biomedgpt,
  title={BiomedGPT: A Unified and Generalist Biomedical Generative Pre-trained Transformer for Vision, Language, and Multimodal Tasks},
  author={Zhang, Kai and Yu, Jun and Yan, Zhiling and Liu, Yixin and Adhikarla, Eashan and Fu, Sunyang and Chen, Xun and Chen, Chen and Zhou, Yuyin and Li, Xiang and others},
  journal={arXiv preprint arXiv:2305.17100},
  year={2023}
}

@article{team2023gemini,
  title={Gemini: a family of highly capable multimodal models},
  author={Team, Gemini and Anil, Rohan and Borgeaud, Sebastian and Wu, Yonghui and Alayrac, Jean-Baptiste and Yu, Jiahui and Soricut, Radu and Schalkwyk, Johan and Dai, Andrew M and Hauth, Anja and others},
  journal={arXiv preprint arXiv:2312.11805},
  year={2023}
}

@article{achiam2023gpt,
  title={Gpt-4 technical report},
  author={Achiam, Josh and Adler, Steven and Agarwal, Sandhini and Ahmad, Lama and Akkaya, Ilge and Aleman, Florencia Leoni and Almeida, Diogo and Altenschmidt, Janko and Altman, Sam and Anadkat, Shyamal and others},
  journal={arXiv preprint arXiv:2303.08774},
  year={2023}
}

@article{alayrac2022flamingo,
  title={Flamingo: a visual language model for few-shot learning},
  author={Alayrac, Jean-Baptiste and Donahue, Jeff and Luc, Pauline and Miech, Antoine and Barr, Iain and Hasson, Yana and Lenc, Karel and Mensch, Arthur and Millican, Katherine and Reynolds, Malcolm and others},
  journal={Advances in Neural Information Processing Systems},
  volume={35},
  pages={23716--23736},
  year={2022}
}

@article{touvron2023llama,
  title={Llama 2: Open foundation and fine-tuned chat models},
  author={Touvron, Hugo and Martin, Louis and Stone, Kevin and Albert, Peter and Almahairi, Amjad and Babaei, Yasmine and Bashlykov, Nikolay and Batra, Soumya and Bhargava, Prajjwal and Bhosale, Shruti and others},
  journal={arXiv preprint arXiv:2307.09288},
  year={2023}
}

@inproceedings{papineni2002bleu,
  title={Bleu: a method for automatic evaluation of machine translation},
  author={Papineni, Kishore and Roukos, Salim and Ward, Todd and Zhu, Wei-Jing},
  booktitle={Proceedings of the 40th annual meeting of the Association for Computational Linguistics},
  pages={311--318},
  year={2002}
}

@inproceedings{banerjee2005meteor,
  title={METEOR: An automatic metric for MT evaluation with improved correlation with human judgments},
  author={Banerjee, Satanjeev and Lavie, Alon},
  booktitle={Proceedings of the acl workshop on intrinsic and extrinsic evaluation measures for machine translation and/or summarization},
  pages={65--72},
  year={2005}
}

@article{romanov2018lessons,
  title={Lessons from natural language inference in the clinical domain},
  author={Romanov, Alexey and Shivade, Chaitanya},
  journal={arXiv preprint arXiv:1808.06752},
  year={2018}
}

@misc{monai,
  doi = {10.5281/ZENODO.4323058},
  url = {https://zenodo.org/record/4323058},
  author = {{MONAI Consortium}},
  title = {MONAI: Medical Open Network for AI},
  publisher = {Zenodo},
  year = {2023},
  copyright = {Apache License 2.0}
}

@misc{mlcommons,
	author = {{MLCommons Consortium}},
	title = {{M}{L}{C}ommons | {M}achine {L}earning {I}nnovation --- mlcommons.org},
	howpublished = {\url{https://mlcommons.org/}},
	year = {},
	note = {[Accessed 07-02-2024]},
}

@inproceedings{delbrouck-etal-2022-vilmedic,
    title = "{V}i{LM}edic: a framework for research at the intersection of vision and language in medical {AI}",
    author = "Delbrouck, Jean-benoit  and
      Saab, Khaled  and
      Varma, Maya  and
      Eyuboglu, Sabri  and
      Chambon, Pierre  and
      Dunnmon, Jared  and
      Zambrano, Juan  and
      Chaudhari, Akshay  and
      Langlotz, Curtis",
    booktitle = "Proceedings of the 60th Annual Meeting of the Association for Computational Linguistics: System Demonstrations",
    month = may,
    year = "2022",
    address = "Dublin, Ireland",
    publisher = "Association for Computational Linguistics",
    url = "https://aclanthology.org/2022.acl-demo.3",
    pages = "23--34",
}

\clearpage
\newpage

\appendix

\setcounter{table}{0}
\renewcommand{\thetable}{A\arabic{table}}

\section{Example of prompts}
\label{appendix:prompts}
We give examples of prompts for each of the datasets that we benchmark in our tool. For each example, we put the text from the dataset in black and the prompt we added in gray.

\begin{table}[htbp]
\footnotesize
\floatconts
  {tab:promptImageClassification}%
  {\caption{Prompt example for each of the image classification datasets.}}%
  {\begin{tabular}{p{0.15\linewidth} | p{0.85\linewidth}}
    \bfseries Dataset & \bfseries Prompt example \\
    \hline \\
    MIMIC-CXR Image Classification & \texttt{\small{\hyphenchar\font=\string"7F $<$img$>$ \color{gray!100}List the conditions that can be seen in this picture.}} \\
    \hline \\
    VinDr Mammo & \texttt{\small{\hyphenchar\font=\string"7F $<$img$>$ \color{gray!100}What is the BI-RADS level in this mammography (from 1 to 5)?}} \\
    \hline \\
    Pad UFES 20 & \texttt{\small{\hyphenchar\font=\string"7F $<$img$>$ Options: Basal Cell Carcinoma (BCC) Squamous Cell Carcinoma (SCC) Actinic Keratosis (ACK) Seborrheic Keratosis (SEK) Bowen’s disease (BOD) Melanoma (MEL) Nevus (NEV) \color{gray!100}What is the most likely diagnosis among the following propositions?}} \\
    \hline \\
    CBIS-DDSM Mass & \texttt{\small{\hyphenchar\font=\string"7F $<$img$>$ \color{gray!100}Is the mass benign, malignant or benign without callback?}} \\
    \hline \\
    CBIS-DDSM Calcification & \texttt{\small{\hyphenchar\font=\string"7F $<$img$>$ \color{gray!100}Is the calcification benign, malignant or benign without callback?}} \\
\end{tabular}}
\end{table}

\begin{table}[htbp]
\footnotesize
\floatconts
  {tab:promptMNIST}%
  {\caption{Prompt example for each of the MedMNIST image classification datasets.}}%
  {\begin{tabular}{p{0.15\linewidth} | p{0.85\linewidth}}
    \bfseries Dataset & \bfseries Prompt example \\
    \hline \\
    OCT MNIST & \texttt{\hyphenchar\font=\string"7F $<$img$>$ Options: \textbackslash n 1: choroidal neovascularization \textbackslash n 2: diabetic macular edema \textbackslash n 3: drusen \textbackslash n 4: normal \textbackslash n \color{gray!100}Which options correspond to the image?} \\
    \hline \\
    Path MNIST & \texttt{\hyphenchar\font=\string"7F $<$img$>$ Options:\textbackslash n 1: adipose \textbackslash n  2: background \textbackslash n  3: debris \textbackslash n  4: lymphocytes \textbackslash n  5: mucus \textbackslash n  6: smooth muscle \textbackslash n  7: normal colon mucosa \textbackslash n  8: cancer-associated stroma \textbackslash n  9: colorectal adenocarcinoma epithelium \textbackslash n  \color{gray!100}Which options correspond to the image?} \\
    \hline \\
    Blood MNIST & \texttt{\hyphenchar\font=\string"7F $<$img$>$ Options:\textbackslash n 1: basophil \textbackslash n  2: eosinophil \textbackslash n  3: erythroblast \textbackslash n  4: immature granulocytes(myelocytes, metamyelocytes and promyelocytes) \textbackslash n  5: lymphocyte \textbackslash n  6: monocyte \textbackslash n  7: neutrophil \textbackslash n  8: platelet \textbackslash n \color{gray!100} Which options correspond to the image?} \\
    \hline \\
    Breast MNIST & \texttt{\small{\hyphenchar\font=\string"7F $<$img$>$ Options:\textbackslash n 1: malignant \textbackslash n  2: normal, benign \textbackslash n  \color{gray!100}Which options correspond to the image?}} \\
    \hline \\
    Derma MNIST & \texttt{\hyphenchar\font=\string"7F $<$img$>$ Options:\textbackslash n 1: actinic keratoses and intraepithelial carcinoma \textbackslash n  2: basal cell carcinoma \textbackslash n  3: benign keratosis-like lesions \textbackslash n  4: dermatofibroma \textbackslash n  5: melanoma \textbackslash n  6: melanocytic nevi \textbackslash n  7: vascular lesions \textbackslash n  \color{gray!100}Which options correspond to the image?} \\
    \hline \\
    OrganC MNIST & \texttt{\hyphenchar\font=\string"7F $<$img$>$ Options:\textbackslash n 1: bladder \textbackslash n  2: femur-left \textbackslash n  3: femur-right \textbackslash n  4: heart \textbackslash n  5: kidney-left \textbackslash n  6: kidney-right \textbackslash n  7: liver \textbackslash n  8: lung-left \textbackslash n  9: lung-right \textbackslash n  10: pancreas \textbackslash n 11: spleen \textbackslash n  \color{gray!100}Which options correspond to the image?} \\
    \hline \\
    OrganS MNIST & \texttt{\hyphenchar\font=\string"7F $<$img$>$ Options:\textbackslash n 1: bladder \textbackslash n  2: femur-left \textbackslash n  3: femur-right \textbackslash n  4: heart \textbackslash n  5: kidney-left \textbackslash n  6: kidney-right \textbackslash n  7: liver \textbackslash n  8: lung-left \textbackslash n  9: lung-right \textbackslash n  10: pancreas \textbackslash n  11: spleen \textbackslash n  \color{gray!100}Which options correspond to the image?} \\
    \hline \\
    Pneumonia MNIST & \texttt{\hyphenchar\font=\string"7F $<$img$>$ Options:\textbackslash n 1: normal \textbackslash n  2: pneumonia \textbackslash n  \color{gray!100}Which options correspond to the image?} \\
    \hline \\
    Retina MNIST & \texttt{\hyphenchar\font=\string"7F $<$img$>$ Options:\textbackslash n 1: 0 \textbackslash n  2: 1 \textbackslash n  3: 2 \textbackslash n  4: 3 \textbackslash n  5: 4 \textbackslash n \color{gray!100} Which options correspond to the image?} \\
    \hline \\
    Tissue MNIST & \texttt{\hyphenchar\font=\string"7F $<$img$>$ Options:\textbackslash n 1: Collecting Duct, Connecting Tubule \textbackslash n  2: Distal Convoluted Tubule \textbackslash n  3: Glomerular endothelial cells \textbackslash n  4: Interstitial endothelial cells \textbackslash n  5: Leukocytes \textbackslash n  6: Podocytes \textbackslash n  7: Proximal Tubule Segments \textbackslash n  8: Thick Ascending Limb \textbackslash n \color{gray!100} Which options correspond to the image?} \\
\end{tabular}}
\end{table}

\begin{table}[htbp]
\footnotesize
\floatconts
  {tab:promptQA}%
  {\caption{Prompt example for each of the QA datasets.}}%
    {\begin{tabular}{p{0.15\linewidth} | p{0.85\linewidth}}
  \bfseries Dataset & \bfseries Prompt example\\
  \hline \\
  MedQA & \texttt{\hyphenchar\font=\string"7F A 67-year-old man with transitional cell carcinoma of the bladder comes to the physician because of a 2-day history of ringing sensation in his ear. He received this first course of neoadjuvant chemotherapy 1 week ago. Pure tone audiometry shows a sensorineural hearing loss of 45 dB. The expected beneficial effect of the drug that caused this patient's symptoms is most likely due to which of the following actions? Options: A: Inhibition of thymidine synthesis. B: Inhibition of proteasome. C: Hyperstabilization of microtubules. D: Generation of free radicals. E: Cross-linking of DNA. \color{gray!100}What is the correct answer?} \\
  \hline \\
  MedMCQA & \texttt{\hyphenchar\font=\string"7F Which of the following is not true for myelinated nerve fibers: a: Impulse through myelinated fibers is slower than non-myelinated fibers. b: Membrane currents are generated at nodes of Ranvier. c: Saltatory conduction of impulses is seen. d: Local anesthesia is effective only when the nerve is not covered by myelin sheath. \color{gray!100}What is the correct answer?}\\
  \hline \\
  PubMedQA & \texttt{\hyphenchar\font=\string"7F \color{gray!100}Answer the question with yes, no or maybe. \color{black}Dyschesia can be provoked by inappropriate defecation movements. The aim of this prospective study was to demonstrate dysfunction of the anal sphincter and/or the musculus (m.) puborectalis in patients with dyschesia using anorectal endosonography. Twenty consecutive patients with a medical history of dyschesia and a control group of 20 healthy subjects underwent linear anorectal endosonography (Toshiba models IUV 5060 and PVL-625 RT). In both groups, the dimensions of the anal sphincter and the m. puborectalis were measured at rest, and during voluntary squeezing and straining. Statistical analysis was performed within and between the two groups. The anal sphincter became paradoxically shorter and/or thicker during straining (versus the resting state) in 85\% of patients but in only 35\% of control subjects. Changes in sphincter length were statistically significantly different (p$<$0.01, chi(2) test) in patients compared with control subjects. The m. puborectalis became paradoxically shorter and/or thicker during straining in 80\% of patients but in only 30\% of controls. Both the changes in length and thickness of the m. puborectalis were significantly different (p$<$0.01, chi(2) test) in patients versus control subjects. Is anorectal endosonography valuable in dyschesia?}\\
  \end{tabular}}
\end{table}

\begin{table}[htbp]
\footnotesize
\floatconts
  {tab:promptVQA}%
  {\caption{Prompt example for each of the VQA datasets.}}%
    {\begin{tabular}{p{0.15\linewidth} | p{0.85\linewidth}}
  \bfseries Dataset & \bfseries Prompt example\\
  \hline \\
  VQA-Rad & \texttt{\hyphenchar\font=\string"7F \color{gray!100}Answer the following question with yes or no. \color{black} $<$img$>$ are regions of the brain infarcted?} \\
  \hline \\
  Path-VQA & \texttt{\hyphenchar\font=\string"7F $<$img$>$ where are liver stem cells (oval cells) located?} \\
  \hline \\
  SLAKE & \texttt{\hyphenchar\font=\string"7F $<$img$>$ What is the main organ in the image?} \\
  \end{tabular}}
\end{table}

\begin{table}[htbp]
\footnotesize
\floatconts
  {tab:promptGeneration}%
  {\caption{Prompt example for report generation and report summarization.}}%
  {\begin{tabular}{p{0.15\linewidth} | p{0.85\linewidth}}
  \bfseries Dataset & \bfseries Prompt example\\
  \hline \\
  MIMIC-III & \texttt{\hyphenchar\font=\string"7F intracranial vessels are all normal in appearance including the carotid arteries and circle of \textunderscore \textunderscore \textunderscore without any aneurysm identified. again noted is the right parafalcine subdural hematoma extending over the right tentorium without significant change from recent prior exam. no gross reaccumulation of the left sided subdural hematoma is appreciated, though evaluation for subtle hemorrhage is difficult as this was a contrast-enhanced study. there has been interval placement of a right sided ventriculostomy tube terminating in the posterior aspect of the right lateral ventricle. again noted is some high attenuation material in the left lateral ventricle, third ventricle and fourth ventricle which likely represents clot adherent to choroid plexus. there is mild ventriculomegaly which is not significantly changed from the prior exam. intraparenchymal high attenuation surrounding the ventriculostomy tube in the right frontal lobe likely represents a small amount of intraparenchymal hemorrhage. the subarachnoid hemorrhage noted in the right sylvian fissure on the prior exam is difficult to appreciate due to contrast-enhancement, but is likely not significantly changed. limited views through the cervical spine demonstrate multilevel spondylosis with mild anterolisthesis of c3 upon c4, likely on the basis of facet degenerative change as no spondylolysis is identified. there is no significant central canal stenosis. incidentally noted is enlargement of the right lobe of the thyroid gland which contains a focus of high attenuation possibly representing calcification. \color{gray!100}Summarize the findings.} \\
  \hline \\
  MIMIC-CXR & \texttt{\hyphenchar\font=\string"7F $<$img$>$ $<$img$>$ $<$img$>$ \color{gray!100}Please caption this scan with findings and impression.}
  \end{tabular}}
\end{table}

\begin{table}[htbp]
\footnotesize
\floatconts
  {tab:promptNLI}%
  {\caption{Prompt example for each of the Natural Language Inference datasets.}}%
  {\begin{tabular}{p{0.15\linewidth} | p{0.85\linewidth}}
    \bfseries Dataset & \bfseries Prompt example \\
    \hline \\
    MedNLI & \texttt{\small{\hyphenchar\font=\string"7F Sentence 1: Labs were notable for Cr 1.7 (baseline 0.5 per old records) and lactate 2.4.\textbackslash n Sentence 2:  Patient has elevated Cr\textbackslash n Determine the logical relationship between these two sentences.\color{gray!100} Does the second sentence logically follow from the first (entailment), contradict the first (contradiction), or if there is no clear logical relationship between them (neutral)?}} \\
\end{tabular}}
\end{table}

\end{document}